\documentclass[10pt,twocolumn,letterpaper]{article}

\usepackage{cvpr}
\usepackage{times}
\usepackage{epsfig}
\usepackage{graphicx}
\usepackage{amsmath}
\usepackage{amssymb}

\usepackage{balance}
\usepackage{multirow}
\usepackage{array}
\newcolumntype{C}[1]{>{\centering\arraybackslash}p{#1}}

\usepackage[pagebackref=true,breaklinks=true,letterpaper=true,colorlinks,bookmarks=false]{hyperref}

\cvprfinalcopy 

\def\cvprPaperID{****} 

\ifcvprfinal\fi
\begin{document}

\title{IA-MOT: Instance-Aware Multi-Object Tracking with Motion Consistency}

\author{
Jiarui Cai, Yizhou Wang, Haotian Zhang, Hung-Min Hsu, Chengqian Ma, Jenq-Neng Hwang\\
Department of Electrical and Computer Engineering\\
University of Washington, Seattle, WA, USA\\
{\tt\small \{jrcai, ywang26, haotiz, hmhsu, cm74, hwang\}@uw.edu}
}

\maketitle

\begin{abstract}
   Multiple object tracking (MOT) is a crucial task in computer vision society. However, most tracking-by-detection MOT methods, with available detected bounding boxes, cannot effectively handle static, slow-moving and fast-moving camera scenarios simultaneously due to ego-motion and frequent occlusion. In this work, we propose a novel tracking framework, called “instance-aware MOT” (IA-MOT), that can track multiple objects in either static or moving cameras by jointly considering the instance-level features and object motions. First, robust appearance features are extracted from a variant of Mask R-CNN detector with an additional embedding head, by sending the given detections as the region proposals. Meanwhile, the spatial attention, which focuses on the foreground within the bounding boxes, is generated from the given instance masks and applied to the extracted embedding features. In the tracking stage, object instance masks are aligned by feature similarity and motion consistency using the Hungarian association algorithm. Moreover, object re-identification (ReID) is incorporated to recover ID switches caused by long-term occlusion or missing detection. Overall, when evaluated on the MOTS20 and KITTI-MOTS dataset, our proposed method won the first place in Track 3 of the BMTT Challenge in CVPR2020 workshops.

\end{abstract}

\section{Introduction}

\begin{figure}[t]
\begin{center}
\includegraphics[width=\linewidth]{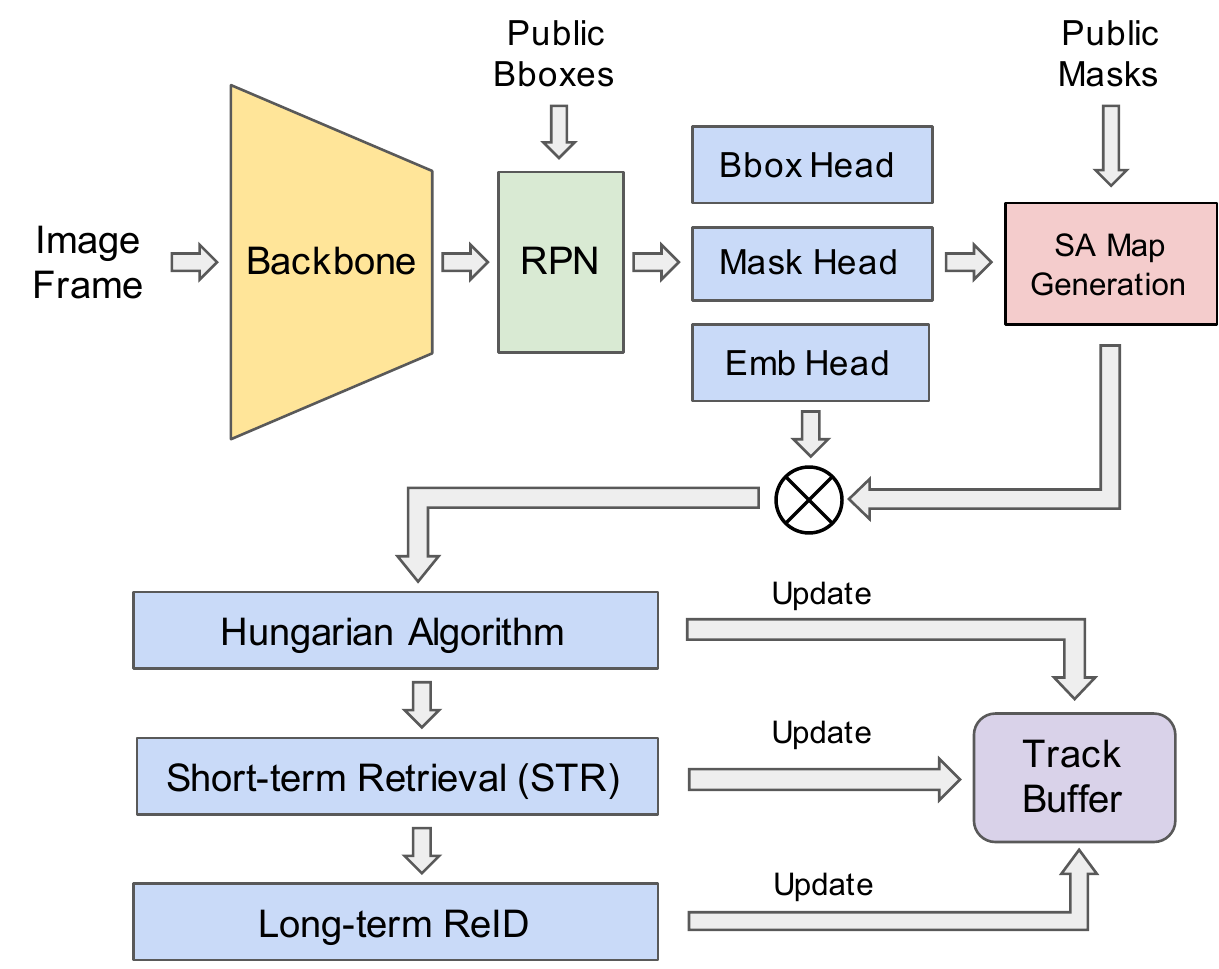}
\end{center}
   \caption{The framework of the proposed IA-MOT. First, embedding features are extracted using a variant of Mask R-CNN and spatial attention generation. Then, isolated detections are aligned by Hungarian algorithm based on the mask IOU and cosine feature similarity. Finally, missing detection and long-term occlusion are handled by the short-term retrieval and ReID modules.}
\label{fig:workflow}
\end{figure}

Multi-Object Tracking (MOT) leveraging detected bounding box locations has been the mainstream and well-studied, a.k.a., tracking-by-detection. Most tracking-by-detection methods firstly extract the objects' deep features from the detected bounding boxes. With these extracted features, a tracking algorithm is followed to associate the detections along consecutive frames. However, the bounding-box-based features contain significant background noises, which are usually redundant and would downgrade the tracking performance. 
Recently, Multi-Object Tracking and Segmentation (MOTS) \cite{voigtlaender2019mots} is posed as a new task to track multiple objects with the corresponding instance segmentation. MOTS is more informative and spatially precise, and the masks also introduce potential promotions to tracking accuracy. 

Typically, there are three key components in MOTS: object detection and segmentation, feature extraction, and multi-object tracking. Some related works are briefly reviewed as follows. Recent studies on object detection and segmentation achieve impressive performance with various advanced structures \cite{ren2015faster,he2017mask}. 
Especially, for mask generation in MOTS, Luiten et al. \cite{luiten2018premvos} present the Proposal-generation, Refinement and Merging for Video Object Segmentation algorithm (PReMVOS). First, a set of object segmentation mask proposals are generated from Mask R-CNN and refined by a refinement network frame-by-frame. Then, with the assistance of optical-flow and ReID information, these selected proposals are merged into pixel-wise object tracks over a video sequence. 
Feature extraction with deep neural networks is widely used in the appearance similarity comparison in tracking. FaceNet \cite{schroff2015facenet} proposes a metric learning strategy. Moreover, a multi-scaled pedestrian detector \cite{jde} is proposed to allow target detection and appearance features to be learned in a shared model.  
On top of that, several MOT methods are studied to obtain the object trajectories. Wang et al. \cite{wang2019exploit} propose the TrackletNet Tracking (TNT) framework that jointly considers appearance and geometric features to form a tracklet-based graph model. 
To incorporate segmentation into MOT, Voigtlaende et al. \cite{voigtlaender2019mots} propose a new baseline approach, Track R-CNN, which addresses detection, tracking, and segmentation via a unified convolutional neural network. Milan et al. \cite{milan2015joint} use superpixel information and consider detection and segmentation jointly with a conditional random field model. CAMOT \cite{ovsep2018track} exploits stereo information to perform mask-based tracking of generic objects on the KITTI dataset. 

Instance segmentation is trained and inferred in parallel with the bounding box detection in the previous studies. In this paper, to better utilize the instance masks for embedding feature extraction, we propose a novel MOTS framework, called instance-aware multi-object tracking (IA-MOT) with motion consistency (IA-MOT) to integrate the segmentation information with instance features. With the given detection results, the sequences are sent into a variant of Mask R-CNN \cite{he2017mask} to obtain the embedding features as region proposals of RPN. Here, to handle instance-aware features, we apply the spatial attention generated from the corresponding instance masks to weight more on the foreground of each bounding box. Second, the extracted instance-aware features are utilized in the following tracking module based on the Hungarian assignment algorithm. In this tracking module, object motion consistency, i.e., the similarity of object motion, including object sizes, moving direction, and moving speed, are jointly considered. Moreover, discontinuity caused by missing detection is recovered by the short-term retrieval (STR) module and object re-identification (ReID) is further incorporated to reduce ID switches caused by long-term occlusion.

\section{The Proposed Method}
The proposed IA-MOT includes three steps: embedding feature extraction, online object tracking with STR, and object re-identification with motion consistency for final refinement. The overall framework is shown in Figure~\ref{fig:workflow}. 

\subsection{Instance-Aware Embedding Features}
First of all, embedding feature for each given bounding box is extracted from a variant of Mask R-CNN \cite{he2016deep} with an extra embedding head. 
This network is trained by the joint optimization of bounding box classification and regression, object mask prediction, and object feature extraction as a multi-task problem. The loss function is
\begin{equation}
  \mathcal{L}_{total} = \alpha_{1} \mathcal{L}_{box} + \alpha_{2} \mathcal{L}_{cls} + \alpha_{3} \mathcal{L}_{mask} + \alpha_{4} \mathcal{L}_{emb},
\end{equation}
where $\mathcal{L}_{box}$, $\mathcal{L}_{cls}$ and $\mathcal{L}_{mask}$ are the original Mask R-CNN loss, $\mathcal{L}_{emb}$ is the cross-entropy loss for object identity classification. Here, each public detection $D_n \in \mathcal{D}$ is treated as a region proposal and fed into the network to acquire its bounding box feature $emb^{box}_n$. 

Then, the extracted embedding features $emb^{box}_n$ are filtered by the spatial attention (SA) map $w_n$ to form our instance-aware (IA) embedding features $emb^{IA}_n$, where 
\begin{equation}
    emb^{IA}_n = w_n \cdot emb^{box}_n.
\end{equation}
Here, the SA map $w_n$ is generated from the provided instance segmentation $mask_n$, where
\begin{equation}
    w_{n}(i,j) = \left\{
    \begin{aligned}
    &1, \ \ \ \ \ \text{if } mask_n(i,j) \text{ is foreground, } \\
    &0.5, \ \ \text{if } mask_n(i,j) \text{ is background. }
    \end{aligned}
    \right.
\end{equation}

\begin{figure*}[t]
\begin{center}
\includegraphics[width=.91\linewidth]{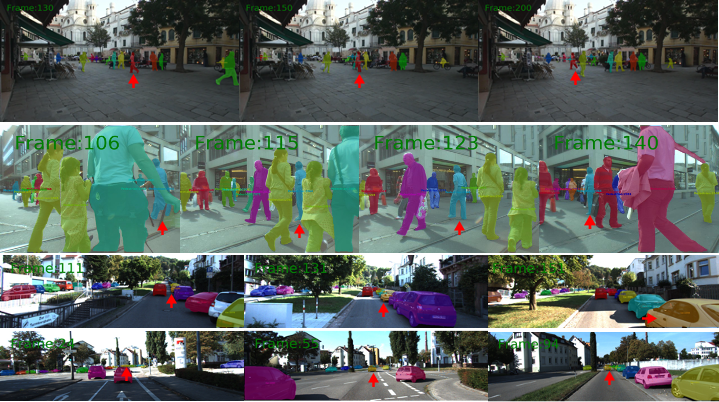}
\end{center}
   \caption{Qualitative results of the proposed IA-MOT. 1$^{\text{st}}$ row: MOTS20-01 (static camera); 2$^{\text{nd}}$ row: MOTS20-06 (stroller-mounted camera); 3$^{\text{rd}}$ row: KITTI-MOTS-0011 (car-mounted camera, up-slope view); 4$^{\text{th}}$ row: KITTI-MOTS-0012 (car-mounted camera, turning view). Red arrows indicate that the targets are tracked robustly even with frequent occlusion or turning.}
\label{fig:example}
\end{figure*}

\subsection{Online Tracking with STR}
After the IA embedding features are extracted, they are transmitted to the following tracking algorithm to associate the same identities among consecutive frames. 
With detections $\mathcal{D}^{(t)}$ and the corresponding feature embeddings $\mathcal{E}^{(t)}$ in frame $t$, the hypotheses ${D_{i}^{(t)} }\in \mathcal{D}^{(t)}$ are matched with live tracks  ${D_{j}^{(t-1)} } \in \mathcal{D}^{(t-1)}$ in frame $t-1$ using the Hungarian assignment algorithm \cite{kuhn1955hungarian}. The assignment cost between the $j$-th track and the $i$-th detection is computed by
\begin{equation}
  \mathcal{C} = 2 - \text{maskIOU}(D_{j}^{(t-1)}, D_{i}^{(t)}) - \text{simi}(emb_{j}, emb_{i}^{(t)}),
\end{equation}
where the $j$-th track's feature $emb_{j}$ is the stack of its first $5$ frames and the most recent $5$ frames, $\text{simi}(\cdot)$ represents the cosine similarity. Considering the visibility of an object may vary due to camera movement, the features are not aggregated into one, but kept separated. Feature similarity is compared pair-wisely, and the maximum is taken as the result. 
Missing detections are compensated by the short-term retrieval (STR) module for lost objects. STR tries to match unassigned detections in frame $t$ with the live tracks that without a detection in frame $t-1$. 

In addition to the feature similarity, the bounding boxes in current frame of the lost track are extrapolated by Huber regression. The distance between the regressed location and the track's last location is confined to within twice the object width. Tracklets will be marked as terminated if there are no alignments for the most recent $N_{1}$ frames and not be included in the Hungarian assignment or STR.

\subsection{ReID with Motion Consistency}
Then, long-term occlusions are recovered by feature-based re-identification (ReID). In this stage, two tracklets $\xi_{u}$ and $\xi_{v}$ without overlapped frames  in time (assuming $\xi_{u}$ is earlier than $\xi_{v}$),  within $N_{2}$ frames apart, and with feature similarity higher than $\beta_{1}$, are considered as possible matched pairs. For static cameras, the interval in between is extrapolated from $\xi_{u}$ and $\xi_{v}$, respectively. $\xi_{u}$ and $\xi_{v}$ are reconnected if the average bounding box IOU between two extrapolation are above $\beta_{2}$. For moving cameras, tracklet motion vectors are estimated from the $\xi_{u}$'s last or $\xi_{v}$'s first $N_{3}$ frames. Thus, the motion of $\xi_u$ can be defined as  
\begin{equation}
\begin{aligned}
  M_{u} &= [M_{ux},M_{uy}] \\
  &= \frac{1}{N_3-1}\left[\sum_{j=1}^{N_{3}-1}{(x^{(j+1)}_{u}-x^{(j)}_{u})},\sum_{j=1}^{N_{3}-1}{(y^{(j+1)}_{u}-y^{(j)}_{u})}\right],
\end{aligned}
\end{equation}
where $[x^{(j)}_u,y^{(j)}_u]$ is the top-left point of the $j$-th detection in $\xi_u$. The motion $M_{v}$ can be defined in the same manner. Then, $\xi_{u}$ and $\xi_{v}$ are recognized as the same object if the cosine similarity between $M_{u}$ and $M_{v}$ are positive and above a threshold $\beta_{3}$.

\section{Experiments}
\subsection{Dataset}
The data of the BMTT Challenge consists of MOTS20 and KITTI-MOTS dataset \cite{voigtlaender2019mots}. There are $8$ sequences in MOTS20 dataset for pedestrian tracking, and evenly split for training and testing. In testing set, the resolution varies from $640 \times 480$ to $1920 \times 1080$ with an average density of $10.6$ targets per frame. KITTI-MOTS is a driving scenario dataset for both pedestrian and car tracking tasks, consisting of $21$ training sequences and $29$ testing sequences. The pre-computed detections are generated from Mask R-CNN X152 \cite{he2017mask} and refined by the refinement net \cite{luiten2018premvos}. 

\subsection{Implementation Details}
The proposed modified Mask R-CNN uses ResNet50 \cite{he2016deep} as the backbone, which is pretrained on COCO dataset \cite{lin2014microsoft} and fine-tuned on the MOTS20 dataset and KITTI-MOTS dataset. Based on the convergence speed and scaling of each loss component, the detection loss weight $\alpha_{1}=\alpha_{2}=1$, mask weight $\alpha_{3}=0.8$ and embedding weight $\alpha_{4}=0.1$. The output feature is $1024$-dim. Short-term memory interval to determine the state of a track is $N_{1}= 0.2$ second, while the long-term interval for ReID is $N_{2}= 1$ second for pedestrian, and $N_{1}= 0.1$ second, $N_{2}= 0.5$ second for cars. $N_{3}= 5$ frames for both categories. 

Moreover, due to the large number of false positives in the provided detections, we create three different filters, including detection confidence, bounding box size, and bounding box aspect ratio, to select valid candidates. After tracking with the filtered detections, short tracks or tracks with low average confidence are discarded. In addition, the trajectory IOU is defined as the average mask IOU of two tracks over their co-exist frames. If the trajectory IOU is larger than $0.75$, the shorter track will be discarded. These processing steps efficiently remove duplicated and non-target detections.

\subsection{MOTS Results}
We evaluate our IA-MOT on Track 3 dataset of the BMTT Challenge (the combination of MOTS20 and KITTI-MOTS) and get the first place, with $69.8$ sMOTSA, out of $13$ methods. The quantitative results are shown in Table~\ref{tab:res} and some qualitative examples are shown in Figure~\ref{fig:example}. 

Specifically, the proposed framework achieves $69.4$ in MOTS20 dataset with the public detections, which is nearly even with the leading method in the MOTS20 leaderboard with private detectors. Although IA-MOT does not get the best performance on KITTI-MOTS, it still ranks the 3$^{\text{rd}}$ for KITTI-MOTS car and 5$^{\text{th}}$ for pedestrian, indicating its ability to generalize to different object categories and the potential to handle different complex MOTS scenarios. 

\begin{table}
\begin{center}
\begin{tabular}{|l|C{0.8cm}|C{0.8cm}|c|c|}
\hline
\multirow{2}{*}{Method} & \multicolumn{2}{c|}{KITTI-MOTS} & \multirow{2}{*}{MOTS20} & \multirow{2}{*}{\underline{Total}} \\
\cline{2-3}
& Car & Ped & & \\
\hline\hline
MCFPA & 77.0 & 67.2 & 66.1 & 69.1 \\
TPM-MOTS & 75.8 & \textbf{67.3} & 66.6 & 69.1 \\
ReMOTS & 72.6 & 64.6 & 67.9 & 68.3  \\
GMPHD\_SAF & 76.2 & 64.3 & 64.3 & 67.3 \\
Lif\_TS & \textbf{77.5} & 55.8 & 65.3 & 66.0 \\
SRF & 71.4 & 60.9 & 60.0 & 63.1 \\
KQD & 74.4 & 61.8 & 57.3 & 62.7 \\
USN & 72.1 & 59.3 & 59.5 & 62.6 \\
YLC & 62.3 & 57.2 & 59.1 & 59.4 \\
SI & 68.5 & 55.5 & 56.2 & 59.1 \\
FK & 64.1 & 54.5 & 54.3 & 56.8 \\
\hline\hline
\textbf{IA-MOT} (Ours) & 76.4 & 64.0 & \textbf{69.4} & \textbf{69.8} \\
\hline
\end{tabular}
\end{center}
\caption{Track 3 evaluation results of the BMTT Challenge, evaluated by sMOTSA. Best results are marked in \textbf{bold}.}
\label{tab:res}
\end{table}

\section{Conclusion}
In this paper, a novel MOTS framework -- IA-MOT is proposed. A Mask R-CNN with an additional embedding head and spatial attention first generate discriminating features. The following MOT stage consists of online Hungarian assignment, short-term retrieve module and ReID. In addition, several implementation details are presented for the MOTS20 and KITTI-MOTS dataset. The proposed framework could effectively track both pedestrian and car, with static or moving cameras, and is flexible for different video resolution and scenarios. Our proposed IA-MOT achieved the winner in Track 3 of the BMTT Challenge in CVPR2020 workshops.   

{\small
\bibliographystyle{ieee_fullname}
\bibliography{egbib}
}

\end{document}